\documentclass[11pt,a4paper]{article}
\usepackage[hyperref]{acl2019}

\aclfinalcopy
\def\aclpaperid{1439}

\usepackage{times}
\usepackage{latexsym}

\usepackage{pifont}
\usepackage{url}

\def\checkmark{\tikz\fill[scale=0.4](0,.35) -- (.25,0) -- (1,.7) -- (.25,.15) -- cycle;}

\usepackage{hyperref}
\usepackage{url}

\usepackage{colortbl}

\usepackage{latexsym}
\usepackage{amssymb}
\usepackage{amsmath}

\usepackage{tikz,tikz-qtree}
\usetikzlibrary{arrows.meta,calc,decorations.markings,math,arrows.meta}
\usepackage{multirow}
\usepackage{xspace} 
\usepackage{subcaption}
\usepackage{graphicx}
\usepackage{array}
\usepackage{tabularx}

\usepackage{csvsimple}
\usepackage{datatool}
\usepackage{booktabs}
\usepackage{subcaption}
\newcommand{\ignore}[1]{}
\usepackage{enumitem}
\usepackage{pgfplots}
\pgfplotsset{compat=1.14}

\usepackage{tikz}
\definecolor{g-red}{HTML}{DB4437}
\definecolor{g-blue}{HTML}{4285F4}
\definecolor{g-green}{HTML}{0F9D58}
\definecolor{g-yellow}{HTML}{F4B400}
\definecolor{g-orange}{HTML}{FF9800}
\definecolor{g-grey}{HTML}{9E9E9E}

\newcommand{\mn}{Full-Transformer\xspace}
\newcommand{\eat}[1]{\ignorespaces}
\newcommand{\mcw}{\mathcal{W}}
\newcommand{\catalog}{dictionary\xspace}
\newcommand{\catalogs}{dictionaries\xspace}

\newcommand{\utgt}{$U_\text{tgt}$\xspace}
\newcommand{\ust}{$U_\text{src$+$tgt}$\xspace}

\newcommand{\suwb}{$U_\text{WB}$\xspace}

\newcommand{\suwbtgt}{$U_\text{WB} \rightarrow U_\text{tgt}$\xspace}
\newcommand{\suwbst}{$U_\text{WB} \rightarrow U_\text{src$+$tgt}$\xspace}
\newcommand{\suwbstt}{$U_\text{WB}\rightarrow U_\text{src$+$tgt} 
\rightarrow U_\text{tgt}$\xspace}

\newcommand{\uwb}{\suwb}

\newcommand{\uwbst}{\suwbst}
\newcommand{\uwbstt}{\suwbstt}

\usepackage[colorinlistoftodos,prependcaption,textsize=tiny]{todonotes}
\newcounter{kt}
\newcommand{\kt}[1]{%
\refstepcounter{kt}%
{%
\todo[color=blue, size=\footnotesize]{%
[\textbf{kt:\thekt}] #1}%
}}%

\newcounter{mw}
\newcommand{\mw}[1]{%
\refstepcounter{mw}%
{%
\todo[color=orange, size=\footnotesize]{%
[\textbf{mw:\thekt}] #1}%
}}%

\newcounter{lj}
\newcommand{\lj}[1]{%
\refstepcounter{lj}%
{%
\todo[color=green, size=\footnotesize]{%
[\textbf{lj:\thekt}] #1}%
}}%
\newcounter{kl}

\newcommand{\PrintDataSimple}[2]{%
\DTLassign{#1}{#2}{%
  \a=d1,%
  \b=d2,%
  \c=d3,%
  \d=d4,%
  \e=d5,%
  \f=d6%
}
\a & \c & \d & \e & \f
}

\newcolumntype{N}{>{\centering\arraybackslash}X}
\newcolumntype{C}[1]{>{\centering\arraybackslash}p{#1}}
\newcolumntype{L}[1]{>{\raggedright\arraybackslash}p{#1}}

\author{
Lajanugen Logeswaran$^\dagger$\thanks{\hspace{.3em} Work completed while interning at Google} \quad 
Ming-Wei Chang$^\ddagger$ \quad 
Kenton Lee$^\ddagger$ \quad 
Kristina Toutanova$^\ddagger$ \\ 
\hspace{-3em} {\bf Jacob Devlin}$^\ddagger$ \qquad 
{\bf Honglak Lee}$^{\ddagger,\dagger}$ \\[.5em]
$^\dagger$University of Michigan, $^\ddagger$Google Research \\
\small{\texttt{\{llajan,honglak\}@umich.edu}}, 
\\\small{\texttt{\{mingweichang,kentonl,kristout,jacobdevlin,honglak\}@google.com}}
}

\begin{document}

\title{Zero-Shot Entity Linking by Reading Entity Descriptions} %

\maketitle

\begin{abstract}
We present the {\em zero-shot entity linking} task, where mentions must be linked to unseen entities without in-domain labeled data. 
The goal is to enable robust transfer to highly specialized domains, and so no metadata or alias tables are assumed. 
In this setting, entities are only identified by text descriptions, and models must rely strictly on language understanding to resolve the new entities. 
First, we show that strong reading comprehension models pre-trained on large unlabeled data can be used to generalize to unseen entities.
Second, we propose a simple and effective adaptive pre-training strategy, which we term \textit{domain-adaptive pre-training} (DAP), to address the domain shift problem associated with linking unseen entities in a new domain.
We present experiments on a new dataset that we construct for this task and show that DAP improves over strong pre-training baselines, including BERT. 
The data and code are available at
{\small \url{https://github.com/lajanugen/zeshel}}.\footnote{zeshel stands for \textbf{ze}ro-\textbf{sh}ot \textbf{e}ntity \textbf{l}inking.}

\end{abstract}

\section{Introduction}
Entity linking systems have achieved high performance in settings where a large set of disambiguated mentions of entities in a target entity \catalog 
is available for training.
Such systems typically use powerful resources such as a high-coverage alias table, structured data, and linking frequency statistics.
For example, \newcite{mw2008learning} show that by only using the prior probability gathered from hyperlink statistics on Wikipedia training articles, one can achieve 90\% accuracy on the task of predicting links in Wikipedia test articles. 

\begin{figure}[t]
\hspace{-.2cm}\includegraphics[width=1.03\columnwidth]{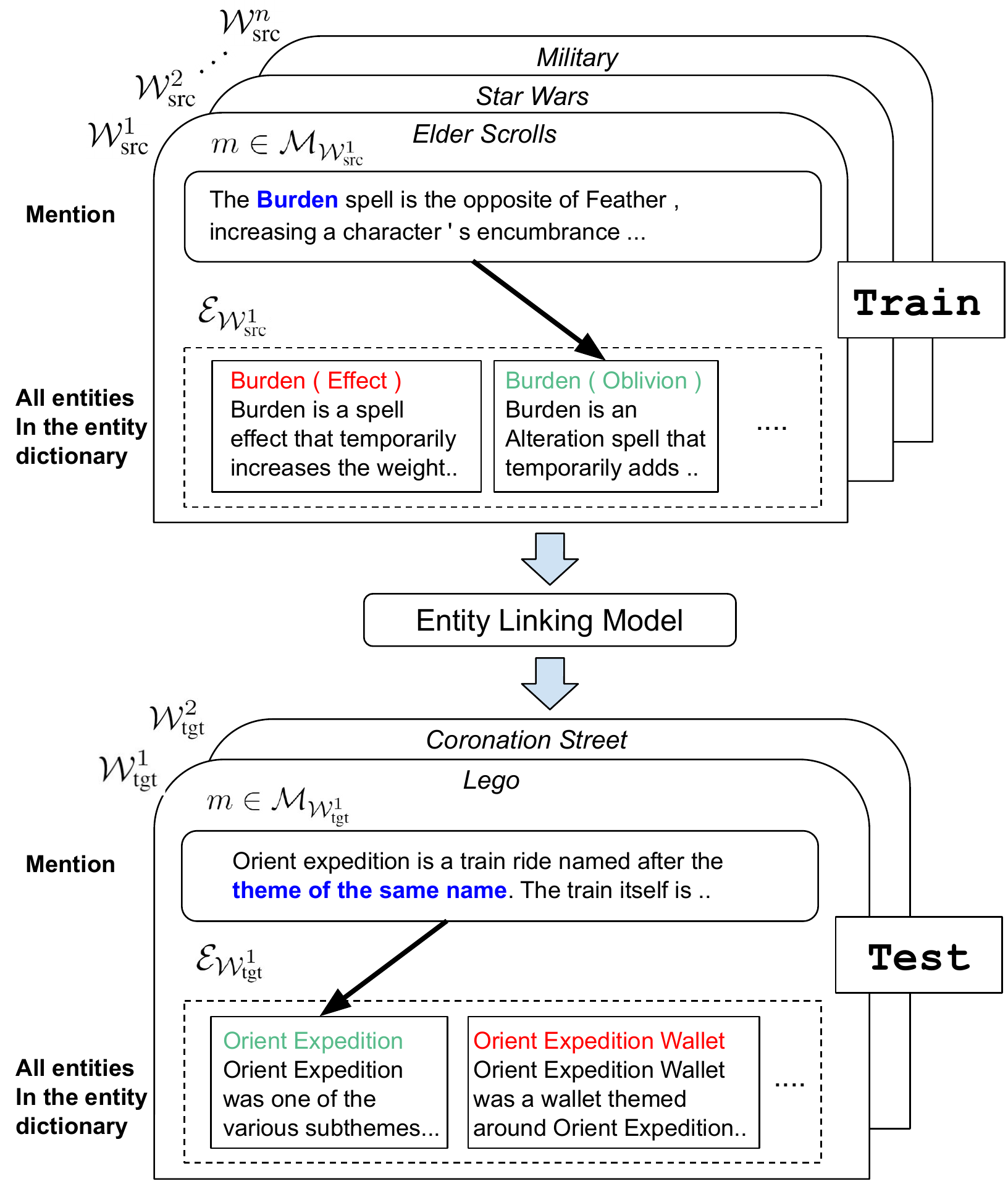}
\caption{
{\bf Zero-shot entity linking.} Multiple training and test domains (worlds) are shown. The
task has two key properties: (1) It is {\bf zero-shot},
as no mentions have been observed for any of the test world entities during training. (2) Only {\bf textual} (non-structured) information is available.
}
\label{fig:example}
\vspace{-1.5em}
\end{figure}

While most prior works focus on linking to  general entity databases, it is often desirable
to link  to specialized entity \catalogs such
as legal cases, company project descriptions, the set of characters in a novel, or a terminology glossary.
Unfortunately, labeled data are not readily available and are often expensive to obtain for these specialized entity \catalogs.
Therefore, we need to develop entity linking systems that can generalize to unseen specialized entities.
Without frequency statistics and meta-data, the task becomes substantially more challenging. Some prior works have pointed out the importance of building entity linking systems that can generalize to unseen entity sets~\cite{anydb,hengji}, but adopt an additional set of assumptions. 

In this work, we propose
a new {\em zero-shot entity linking} task, and construct a new dataset for it.\footnote{Existing datasets are either unsuitable or would have to be artificially partitioned to construct a dataset for this task.}
The target \catalog is simply defined as a set of entities, each with a text description (from a canonical entity page, for example).
We do not constrain mentions to named entities, unlike some prior work, which makes the task harder due to large number of candidate entities.
In our dataset, multiple entity \catalogs are available for training, with task performance measured on a disjoint set of test entity \catalogs for which no labeled data is available. Figure~\ref{fig:example} illustrates the task setup.
We construct the dataset using multiple sub-domains in Wikia and automatically extract labeled mentions using hyper-links.

Zero-shot entity linking poses two challenges for entity linking models.
First, without the availability of powerful alias tables or frequency priors, models must read entity descriptions and reason about the correspondence with the mention in context. We show that a strong reading comprehension model is crucial. Second, since labeled mentions for test entities are not available, models must adapt to new mention contexts and entity descriptions. 
We focus on both of these challenges.

The contributions of this paper are as follows:
\begin{itemize}[itemsep=-0.5ex,leftmargin=*,topsep=2pt]
    \item We propose a new {\em zero-shot entity linking} task that aims to challenge the generalization ability of entity linking systems with minimal assumptions. We construct a dataset for this task, which will be made publicly available. 
    
    \item We build a strong baseline by using state-of-the-art reading comprehension models. We show that  attention between mention in context and entity descriptions, which has not been used in prior entity linking work, is critical for this task. 
    
    \item We propose a simple yet novel adaptation strategy called domain-adaptive pre-training (DAP) and show that it can further improve  entity linking performance.

\end{itemize}

\newcommand\yesmark{\ding{51}}
\begin{table*}[!t]
\centering
\small
\scalebox{1.0}{
\begin{tabular}{lcccccc}
\toprule 
\multirow{2}{*}{Task} & \multirow{2}{*}{In-Domain} & Seen & Small & \multirow{2}{*}{Statistics} & Structured &  Entity \\
 & & Entity Set & Candidate Set & & Data & \catalog \\
\midrule
Standard EL & \yesmark  & \yesmark  &  &  \yesmark &  \yesmark  & \yesmark \\ 
Cross-Domain EL & & \yesmark &  & \yesmark  &  \yesmark  & \yesmark\\ %
Linking to Any DB\scriptsize~\cite{anydb}& & & \yesmark & & \yesmark & \yesmark \\ \cmidrule{1-7}
Zero-Shot EL  & & & & & & \yesmark \\ %
\bottomrule
\end{tabular}
}
\caption{
Assumptions and resources for entity linking task definitions.
We classify task definitions based on whether (i) the system is tested on mentions from the training domain (In-Domain), (ii) linked mentions from the target entity set are seen during training (Seen Entity Set), (iii) a small high-coverage candidate set can be derived using alias tables or strict token overlap constraints (Small Candidate Set) and the availability of (iv) Frequency statistics, (v) Structured Data, and (vi) textual descriptions (Entity \catalog). 
} 
\label{tab:zeshel_compare}
\end{table*}

\section{Zero-shot Entity Linking}
\label{sec:taskdef}

We first review standard entity linking task definitions and discuss assumptions made by prior systems. 
We then define the zero-shot entity linking task and discuss its relationship to prior work.

\subsection{Review: Entity linking}
Entity linking (EL) is the task of grounding entity mentions by linking them to entries in a given database or \catalog of entities. Formally, given a mention $m$ and its context, an entity linking system links $m$ to the corresponding entity in an {\bf entity set} $\mathcal{E} = \{e_i\}_{i=1,...,K}$, where $K$ is the number of entities.
The standard definition of EL \cite{bunescu2006using, eltutorial14,elmultitutorial} assumes that mention boundaries are provided  by users or
a mention detection system. The entity set $\mathcal{E}$ can contain tens of thousands or even millions of entities, making this a challenging task. In practice, many entity linking systems rely on the following resources or assumptions:

\vspace{-2pt}
\paragraph{Single entity set}
This assumes that there is a single comprehensive set of entities $\mathcal{E}$ shared between training and test examples.

\vspace{-2pt}
\paragraph{Alias table}
An alias table contains entity candidates for a given mention string and limits the possibilities to a relatively small set.
Such tables are often compiled from a labeled training set and domain-specific heuristics.

\vspace{-2pt}
\paragraph{Frequency statistics}
Many systems use frequency statistics obtained from a large labeled corpus to estimate entity popularity and the probability of a mention string linking to an entity.
These statistics are very powerful when available. 

\vspace{-2pt}
\paragraph{Structured data}
Some systems assume access to structured data such as relationship tuples (e.g., ({\textit{Barack Obama}, \textit{Spouse}, \textit{Michelle Obama}})) or a type hierarchy to aid disambiguation. 

\subsection{Task Definition}

The main motivation for  this task is to expand the scope of entity linking systems
and make them generalizable to unseen entity sets for which none of the powerful resources listed above are readily available. 
Therefore, we drop the above assumptions and make one weak assumption: the existence of an {\bf entity \catalog} $\mathcal{E} = \{ (e_i, d_i) \}_{i=1,..,K}$, where
$d_i$ is a text description of entity $e_i$.

Our goal is to build entity linking systems that can generalize to new domains and entity \catalogs, which we term \emph{worlds}.
We define a world as $\mcw = (\mathcal{M}_\mcw, \mathcal{U}_\mcw, \mathcal{E}_\mcw)$,
where $\mathcal{M}_\mcw$ and $\mathcal{U}_\mcw$ are distributions over mentions and documents from the world, respectively, and $\mathcal{E}_\mcw$ is an entity \catalog associated with $\mcw$.
Mentions $m$ from $\mathcal{M}_\mcw$ are defined as mention spans in documents from $\mathcal{U}_\mcw$. %
We assume the availability of labelled mention, entity pairs from one or more source worlds  $\mcw_\text{src}^1 \ldots \mcw_\text{src}^n$ for training.
At test time we need to be able to label mentions in a new world $\mcw_\text{tgt}$. 
Note that the entity sets $\mathcal{E}_{\mcw_\text{src}^1}, \ldots, \mathcal{E}_{\mcw_\text{src}^n}, \mathcal{E}_{\mcw_\text{tgt}}$ are disjoint. See Figure \ref{fig:example} for an illustration of several training and test worlds.

We additionally assume that samples from the document distribution $\mathcal{U}_{\mcw_\text{tgt}}$ and the entity descriptions $\mathcal{E}_{\mcw_\text{tgt}}$ are  available for training. 
These samples can be used for unsupervised adaptation to the target world. During training, mention boundaries for mentions in  $\mcw_\text{tgt}$ are not available. At test time, mention boundaries are provided as input.  

\eat{
\kt{Do we need to talk about models here? This is the task definition}
We consider models which learn a scoring function $f(m, e_d)$ which measures the semantic compatibility between mention  $m$ in context, and entity description $e_d$.
Predictions at test time are made by identifying the correct entity from all possible entities in the target world entity set: $\argmax_{(e, e_d)\in \mathcal{E}_{\mcw_\text{tgt}}} f(m, e_d).$
}

\eat{
\subsection{Discussion}

The  Zero-Shot Entity linking task
has two interesting properties. First, because only textual descriptions are available for the target entities $\mathcal{E}_{\mcw_\text{tgt}}$, 
 systems for this task  behave like  {\em reading comprehension} systems, which estimate the semantic compatibility of the mention in context and the candidate entity descriptions. 
Second, unlike common out-of-domain entity linking tasks, in addition to the distribution of document texts $\mathcal{U}$, the set of possible entities shifts entirely from the training distribution. 
Labeled data from multiple training worlds, and unlabeled text from test worlds are available, offering the opportunity to study multi-source unsupervised domain adaptation methods. 
}

\subsection{Relationship to other EL tasks}

We summarize the relationship between the newly introduced zero-shot entity linking task and prior EL task definitions in Table \ref{tab:zeshel_compare}. 
\paragraph{Standard EL} While there are numerous differences between EL datasets \cite{bunescu2006using, Q15-1023}, most focus on a standard setting where mentions from a comprehensive test entity \catalog (often Wikipedia) are seen during training, and rich statistics and meta-data can be utilized \cite{eltutorial14}. Labeled in-domain documents with mentions are also assumed to be available.
\paragraph{Cross-Domain EL} Recent work has also generalized to a cross-domain setting,  linking entity mentions in different types of text, such as blogposts and news articles to the Wikipedia KB, while only using labeled mentions in Wikipedia for training (e.g., ~\citet{unseenroth, titov18}, \textit{inter alia}). \paragraph{Linking to Any DB}\newcite{anydb} proposed a task setup very similar to ours, and later work \cite{hengji} has followed a similar setting. The main difference between zero-shot EL and these works is that they assumed either a high-coverage alias table or high-precision token overlap heuristics to reduce the size of the entity candidate set (i.e., to less than four in \newcite{anydb}) and relied on structured data to help disambiguation. 
By compiling and releasing a multi-world dataset focused on learning from textual information, we hope to help drive progress in linking entities for a broader set of applications.

Work on word sense disambiguation based on dictionary definitions of words is related as well~\cite{wsd18}, but this task exhibits lower ambiguity and existing  formulations have not focused on domain generalization.

\section{Dataset Construction}

We construct a new dataset to study the zero-shot entity linking problem using documents from Wikia.\footnote{ \url{https://www.wikia.com}.}
Wikias are community-written encyclopedias, each specializing in a particular subject or theme such as a fictional universe from a book or film series. 
Wikias have many interesting properties suitable for our task.
Labeled mentions can be automatically extracted based on hyperlinks.
Mentions and entities have rich document context that can be exploited by reading comprehension approaches. 
Each Wikia has a large number of unique entities relevant to a specific theme, making it a useful benchmark for evaluating domain generalization of entity linking systems.

We use data from 16 Wikias, and use 8 of them for training and 4  each for validation and testing.
To construct data for training and evaluation, we first extract a large number of mentions from the Wikias.
Many of these mentions can be easily linked by string matching between mention string and the title of entity documents.
These mentions are downsampled during dataset construction, and occupy a small percentage (5\%) of the final dataset. 
While not completely representative of the natural distribution of mentions, this  data construction method follows recent work that focuses on evaluating performance on the challenging aspects of the entity linking problem (e.g., \newcite{unseenroth} selected mentions with multiple possible entity candidates for assessing in-domain unseen entity performance).
Each Wikia document corresponds to an entity, represented by the title and contents of the document.
These entities, paired with their text descriptions, comprise the entity dictionary.

Since the task is already quite challenging, we assume that the target entity exists in the entity \catalog and leave NIL recognition or clustering (NIL mentions/entities refer to entities non-existent in the knowledge-base) to future editions of the task and dataset.

\begin{filecontents*}{data/zeshel_data_v2.csv}
Set,Description,Num Documents,Num Mentions,,
Train,,,,,
American Football,"The sport of American Football. Teams, players, etc.",31929,3898,410,333
Doctor Who,The long running British Sci-Fi show.,40281,8334,819,702
Fallout,The Fallout video game universe. Post apocalpypic RPG.,16992,3286,337,256
Final Fantasy,"The Final Fantasy video game universe, a set of Japanese RPGs.",14044,6041,629,527
Military,Real-world military history.,104520,13063,1356,1408
Pro Wrestling,"Professional wresling, i.e., WWE and others.",10133,1392,151,111
StarWars,The StarWars universe. The movies/TV shows/books/etc.,87056,11824,1143,1563
World of Warcraft,The popular massively multiplayer online RPG.,27677,1437,155,100
Val,,,,,
Coronation Street,A long running British Soap Opera.,17809,0,0,1464
Muppets,"The Muppets (i.e., Sesame Street) TV shows/movies/etc.",21344,0,0,2028
Ice Hockey,The sport of Ice Hockey.,28684,0,0,2233
Elder Scrolls,The Elder Scrolls video game universe. Medieval fantasy RPG.,21712,0,0,4275
Test,,,,,
Forgotten Realms,A campaign setting/fictional universe for Dungeons and Dragons.,15603,0,0,1200
Lego,The universe around Lego toys. Movies/video games/Lego sets.,10076,0,0,1199
Star Trek,The Star Trek sci-fi universe. TV show/movies/etc.,34430,0,0,4227
YuGiOh,A Japanese Anime and trading card game.,10031,0,0,3374
\end{filecontents*}
\DTLloaddb[noheader,keys={d1,d2,d3,d4,d5,d6}]{data}{data/zeshel_data_v2.csv}
\begin{table}[!t]
\small
\centering
\begin{tabular}{l r r r r}
\toprule
World & Entities & \multicolumn{3}{c}{Mentions} \\\cline{3-5}
\rule{0pt}{2ex}
& & Train & \multicolumn{2}{c}{Evaluation} \\
& &   & Seen & Unseen \\
\midrule
\multicolumn{5}{c}{\textbf{Training}} \\
\midrule
\PrintDataSimple{data}{3} \\
\PrintDataSimple{data}{4} \\
\PrintDataSimple{data}{5} \\
\PrintDataSimple{data}{6} \\
\PrintDataSimple{data}{7} \\
\PrintDataSimple{data}{8} \\
\PrintDataSimple{data}{9} \\
\PrintDataSimple{data}{10} \\
\midrule
\multicolumn{5}{c}{\textbf{Validation}} \\
\midrule
\PrintDataSimple{data}{12} \\
\PrintDataSimple{data}{13} \\
\PrintDataSimple{data}{14} \\
\PrintDataSimple{data}{15} \\
\midrule
\multicolumn{5}{c}{\textbf{Test}} \\
\midrule
\PrintDataSimple{data}{17} \\
\PrintDataSimple{data}{18} \\
\PrintDataSimple{data}{19} \\
\PrintDataSimple{data}{20} \\
\bottomrule
\end{tabular}
\caption{\small Zero-shot entity linking dataset based on Wikia.}
\label{wikiastats}
\vspace{-0.6em}
\end{table}

\begin{table}[t]
\centering
\footnotesize
\scalebox{.91}{
\begin{tabular}{c|p{1.3cm}p{4.3cm}}
\toprule
\multicolumn{3}{c}{\bf Coronation Street} \\
\midrule
Mention & \multicolumn{2}{p{6cm}}{
She told ray that Dickie and Audrey had met up again and tried to give their marriage another go  \ldots I don't want to see {\bf her} face again \ldots "}\\
\midrule 
 & Dickie Fleming & 
Richard ``Dickie" Fleming lived in coronation street with his wife Audrey from 1968 to 1970. \\
\checkmark & Audrey Fleming & Audrey Fleming (ne\'e bright) was a resident of 3 coronation street from 1968 to 1970 . 
Audrey married Dickie Fleming \ldots \\
 & Zeedan Nazir & 
Zeedan Nazir is the son of the Late Kal and Jamila Nazir \ldots  \\
\midrule 
\multicolumn{3}{c}{\bf Star Wars} \\
\midrule
Mention & \multicolumn{2}{p{6cm}}{
The droid acted as Moff Kilran's representative on board the Black Talon, an {\bf Imperial transport ship}. }\\
\midrule 
\checkmark & Gage-class transport &
The Gage-class transport was a transport design used by the reconstituted Sith Empire of the Great Galactic War. \\
& {Imperial Armored Transport}&
The Kuat Drive Yards Imperial Armored Transport was fifty meters long and carried ten crewmen and twenty soldiers. \\ 
& M-class Imperial Attack Transport &
The M-class Imperial Attack Transport was a type of starship which saw service in the Imperial Military during the Galactic War. \\
\bottomrule 
\end{tabular}
}
\caption{Example mention and entity candidates from Coronation Street and Star Wars. Note that the language usage is very different across different Worlds.}
\label{example}
\vspace{-0.5em}
\end{table}

We categorize the mentions based on token overlap between mentions and the corresponding entity title as follows.
\textit{High Overlap}: title is identical to mention text,
\textit{Multiple Categories}: title is mention text followed by a disambiguation phrase (e.g., mention string: `Batman', title: `Batman (Lego)'), 
\textit{Ambiguous substring}: mention is a substring of title (e.g., mention string: `Agent', title: `The Agent').
All other mentions are categorized as \textit{Low Overlap}.
These mentions respectively constitute approximately 5\%, 28\%, 8\% and 59\% of the mentions in the dataset.

Table \ref{wikiastats} shows some statistics of the dataset.
Each domain has a large number of entities ranging from 10,000 to 100,000.
The training set has 49,275 labeled mentions.
To examine the in-domain generalization performance, we construct heldout sets \textit{seen} and \textit{unseen} of 5,000 mentions each, composed of mentions that link to only entities that were seen or unseen during training, respectively.
The validation and test sets have 10,000 mentions each (all of which are unseen).

Table \ref{example} shows examples of mentions and entities in the dataset.
The vocabulary and language used in mentions and entity descriptions differs drastically between the different domains.
In addition to acquiring domain specific knowledge, understanding entity descriptions and performing reasoning is required in order to resolve mentions.

\section{Models for Entity Linking}
We adopt a two-stage pipeline consisting of a fast candidate generation stage, followed by a more expensive but powerful candidate ranking stage.

\subsection{Candidate generation}
Without alias tables for standard entity linking, a natural substitute is to use an IR approach for candidate generation.
We use BM25, a variant of TF-IDF to measure similarity between mention string and candidate documents.\footnote{We also experimented with using the mention+context text but this variant performs substantially worse.}
Top-$k$ entities retrieved by BM25 scoring with Lucene\footnote{~\url{http://lucene.apache.org/}} are used for training and evaluation.
In our experiments $k$ is set to 64. 
The coverage of the top-64 candidates is less than 77\% on average, indicating the difficulty of the task and leaving substantial room for improvement in the candidate generation phase.

\subsection{Candidate ranking}

Since comparing two texts---a mention in context and a candidate entity description---is a task similar to reading comprehension and natural language inference tasks, we use an architecture based on a deep Transformer~\cite{vaswani-etal:2017:_atten} which has achieved state-of-the-art performance on such tasks ~\cite{radford-etal:2018,devlin2018bert}.

As in BERT~\cite{devlin2018bert}, the mention in context $m$ and candidate entity description $e$, each represented by 128 word-piece tokens, are concatenated and input to the model as a sequence pair together with special start and separator tokens: $([\texttt{{CLS}}]\text{ }m \text{ } [\texttt{{SEP}}] \text{ } e \text{ } [\texttt{{SEP}}])$.
Mention words are signaled by a special embedding vector that is added to the mention word embeddings.
The Transformer encoder produces a vector representation $h_{m, e}$ of the input pair, which is the output of the last hidden layer at the special pooling token $[\texttt{{CLS}}]$.
Entities in a given candidate set are scored as $w^\top h_{m, e}$ where $w$ is a learned parameter vector, and the model is trained using a softmax loss.
An architecture with 12 layers, hidden dimension size 768 and 12 attention heads was used in our experiments. 
We refer to this model as \textbf{\mn}. %
By jointly encoding the entity description and the mention in context with a Transformer, they can attend to each other at every layer.

Note that prior neural approaches for entity linking have not explored such architectures with deep cross-attention. To assess the value of this departure from prior work, we implement the following two variants:
{(\textit{i})} \textbf{Pool-Transformer}: a siamese-like network which uses two deep Transformers to separately derive single-vector representations of the mention in context, $h_m$, and the candidate entity, $h_e$; they take as input the mention in context and entity description respectively, together with special tokens indicating the boundaries of the texts: $([\texttt{{CLS}}]\text{ }m \text{ } [\texttt{{SEP}}])$ and $([\texttt{{CLS}}]\text{ }e \text{ } [\texttt{{SEP}}])$, and output the last hidden layer encoding at the special start token. The scoring function is $h_m ^\top h_e$. Single vector representations for the two components have been used in many prior works, e.g., \newcite{unseenroth}.
{(\textit{ii})} \textbf{Cand-Pool-Transformer}: a variant which uses single vector entity representations but can attend to individual tokens of the mention and its context as in \newcite{ganea-hofmann:2017:EMNLP2017}. This architecture also uses two Transformer encoders, but introduces an additional attention module which allows $h_e$ to attend to individual token representations of the mention in context.

In the experiments section, we also compare to re-implementations of \newcite{unseenroth} and \newcite{ganea-hofmann:2017:EMNLP2017}, which are similar to Pool-Transformer and Cand-Pool-Transformer respectively but with different neural architectures for encoding.

\section{Adapting to the Target World}
\label{sec:adapt}
We focus on using unsupervised pre-training to ensure that downstream models are robust to target domain data. 
There exist two general strategies for pre-training: (1) task-adaptive pre-training, and (2) open-corpus pre-training. We describe these below, and also propose a new strategy: domain-adaptive pre-training (DAP), which is complementary to the two existing approaches.

\paragraph{Task-adaptive pre-training} \newcite{glorot2011domain,chen2012marginalized,yang-eisenstein:2015:NAACL-HLT}, \textit{inter alia}, pre-trained on the source and target domain unlabeled data jointly with the goal of discovering features that generalize across domains. After pre-training, the model is fine-tuned on the source-domain labeled data.\footnote{In many works, the learned representations are kept fixed and only higher layers are updated.}

\paragraph{Open-corpus pre-training} Instead of explicitly adapting to a target domain, this approach simply applies unsupervised pre-training to large corpora before fine-tuning on the source-domain labeled data. 
Examples of this approach include ELMo~\cite{peters-etal:2018:_deep}, OpenAI GPT~\cite{radford-etal:2018}, and BERT~\cite{devlin2018bert}. Intuitively, the target-domain distribution is likely to be partially captured by pre-training if the open corpus is sufficiently large and diverse. Indeed, open-corpus pre-training has been shown to benefit out-of-domain performance far more than in-domain performance~\cite{he2018jointly}. 
\paragraph{Domain-adaptive pre-training} In addition to pre-training stages from other approaches, we propose to insert a penultimate \emph{domain adaptive pre-training} (DAP) stage, where the model is pre-trained \emph{only} on the target-domain data. As usual, DAP is followed by a final fine-tuning stage on the source-domain labeled data. The intuition for DAP is that representational capacity is limited, so models should prioritize the quality of target domain representations above all else.
\\\\
We introduce notation to describe various ways in which pre-training stages can be composed.
\begin{itemize}
    \item $U_\text{src}$ denotes text segments from the union of source world document distributions $\mathcal{U}_{\mcw_\text{src}^1} \ldots \mathcal{U}_{\mcw_\text{src}^n}$.  
    \item $U_\text{tgt}$ denotes text segments from the document distribution of a target world $\mathcal{W}_\text{tgt}$.  
    \item \ust denotes randomly interleaved text segments from both $U_\text{src}$ and $U_\text{tgt}$.
    \item \uwb denotes text segments from open corpora, which in our experiments are Wikipedia and the BookCorpus datasets used in BERT.
\end{itemize}

We can chain together a series of pre-training stages. For example, \uwbstt indicates that the model is first pre-trained on the open corpus, then pre-trained on the combined source and target domains, then pre-trained on only the target domain, and finally fine-tuned on the source-domain labeled data.\footnote{We use the notation $U_x$ interchangeably to mean both the unsupervised data $x$ and the strategy to pre-train on $x$.}
We show that chaining together different pre-training strategies provides additive gains.

\eat{
In addition to the above, we also consider multi-stage pre-training strategies where a model is sequentially pre-trained on different subsets of the data.
\begin{itemize}[leftmargin=*]
\item \textbf{(Src + Tgt) followed by Tgt} First pre-train in \textbf{Src + Tgt} setting and further pre-train in \textbf{Tgt only} setting.
\end{itemize}
Furthermore, we consider initializing models with BERT (pre-trained on Wikipedia) to observe the effect of pre-training on a large corpus.

Models are pre-trained on Wikia documents.
Text spans appearing next to each other in the documents are used in paired-sequence mode for training.
\lj{Review comment: Not clear at all}
We pre-train using the Mask-LM objective in \citep{devlin2018bert}.

\mw{It is very important to have the faithful implementation of previous work system
such that we can add citation to the table.}

}

\section{Experiments}

\paragraph{Pre-training} 
We use the BERT-Base model architecture in all our experiments.
The Masked LM objective \cite{devlin2018bert} is used for unsupervised pre-training.
For fine-tuning language models (in the case of multi-stage pre-training) and fine-tuning on the Entity-Linking task, we use a small learning rate of 2e-5, following the recommendations from \citet{devlin2018bert}.
For models trained from scratch we use a learning rate of 1e-4.

\paragraph{Evaluation} We define the \textit{normalized} entity-linking performance as the performance evaluated on the subset of test instances for which the gold entity is among the top-k candidates retrieved during candidate generation.
The \textit{unnormalized} performance is computed on the entire test set.
Our IR-based candidate generation has a top-64 recall of 76\% and 68\% on the validation and test sets, respectively.
The unnormalized performance is thus upper-bounded by these numbers.
Strengthening the candidate generation stage improves the unnormalized performance, but this is outside the scope of our work.
Average performance across a set of worlds is computed by macro-averaging.
Performance is defined as the accuracy of the single-best identified entity (top-1 accuracy).

\subsection{Baselines}

\begin{table}[t]
\centering
\small
\begin{tabularx}{\linewidth}{lNC{1.15cm}}
\toprule
Model & Resources & Avg Acc \\
\midrule
\multicolumn{1}{l}{Edit-distance}   & $\emptyset$ & 16.49 \\ 
\multicolumn{1}{l}{TF-IDF \footnote{We used BM25.}} & $\emptyset$ & 26.06 \\
\multicolumn{1}{l}{\footnotesize \citet{ganea-hofmann:2017:EMNLP2017}} & GloVe & 26.96 \\
\multicolumn{1}{l}{\footnotesize \citet{unseenroth}} & GloVe & 27.03 \\ \midrule
\mn & $\emptyset$ & 19.17\\
\mn (Pre-trained) & $U_\text{src}$ & 66.55 \\
\mn (Pre-trained) & $U_\text{tgt}$ & 67.87 \\
\mn (Pre-trained) & \ust & 67.91\\
\midrule
Pool-Transformer (Pre-trained)& \uwb & 57.61 \\
Cand-Pool-Trans. (Pre-trained) & \uwb & 52.62 \\
\mn (Pre-trained)& \uwb & 76.06 \\
\bottomrule
\end{tabularx}
\caption{Baseline results for Zero-shot Entity Linking. Averaged normalized Entity-Linking accuracy on all validation domains. \ust refers to masked language model pre-training on unlabeled data from training and validation worlds.
}
\label{tab:baseline}
\end{table}

We first examine some baselines for zero-shot entity linking in Table~\ref{tab:baseline}.
We include naive baselines such as Levenshtein edit-distance and TF-IDF, which compare the mention string against candidate entity title and full document description, respectively, to rank candidate entities.

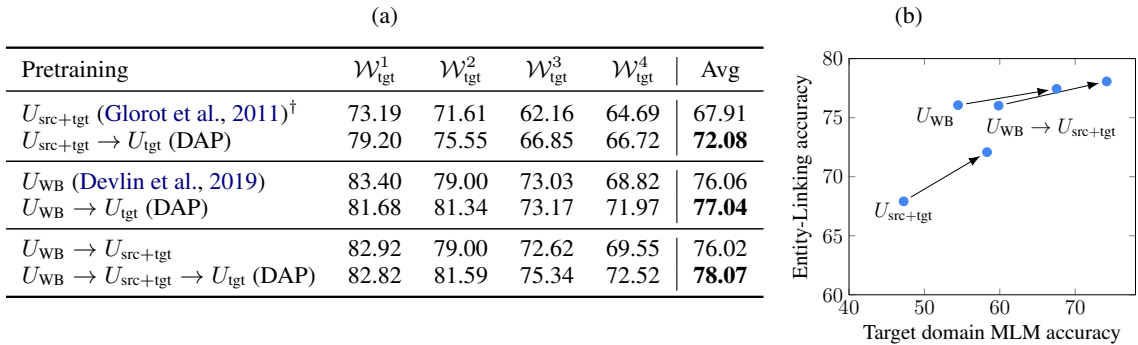
\begin{figure*}[!th]
\hspace{-.3cm}
\begin{subfigure}[t]{0.65\textwidth}
\centering
\small
\caption{}
\label{table:multipretrain}
\begin{tabular}{lcccc|c}
\toprule
Pretraining &$\mcw_\text{tgt}^1$ &$\mcw_\text{tgt}^2$
&$\mcw_\text{tgt}^3$
&$\mcw_\text{tgt}^4$&Avg\\
\midrule
\ust {\footnotesize~\cite{glorot2011domain}$^{\dagger}$}   & 73.19 & 71.61 & 62.16	& 64.69	& 67.91 \\
\ust $\rightarrow$ \utgt (DAP)                             & 79.20 & 75.55 & 66.85	& 66.72	& {\bf 72.08} \\
\midrule 
\suwb{\footnotesize~\cite{devlin2018bert}}                 & 83.40 & 79.00 & 73.03	& 68.82	& 76.06 \\
\suwbtgt (DAP)                                             & 81.68 & 81.34 & 73.17	& 71.97	& {\bf 77.04} \\
\midrule
\suwbst                                                    & 82.92 & 79.00 & 72.62	& 69.55	& 76.02 \\
\suwbstt (DAP)                                             & 82.82 & 81.59 & 75.34	& 72.52	& {\bf 78.07} \\
\bottomrule
\end{tabular}
\end{subfigure}
\begin{subfigure}[t]{0.2\textwidth}
\caption{}
\label{fig:MLM_EL}
\begin{tikzpicture}[scale=0.55]
\begin{axis}[
  nodes near coords align=right, %
  xmax=78,
  xmin=40,
  ymax=80,
  ymin=60,
  xlabel={Target domain MLM accuracy},
  ylabel={Entity-Linking accuracy},
  legend style={nodes={scale=0.8, transform shape}}, 
  legend pos=south east,
  mark size=3pt,
  font=\Large
]

\tikzstyle{atp} = [-{Latex[scale=1.5]}, shorten >=0.2cm,shorten <=0.2cm]
\coordinate (src_tgt) at (47.25, 67.91);
\coordinate (src_tgt_atp) at (58.3075, 72.08);
\draw [atp] (src_tgt) -- (src_tgt_atp);
\node at (47.25, 67.91-1) {\ust};

\coordinate (wb) at (54.4625, 76.06);
\coordinate (wb_atp) at (67.54, 77.44);
\draw [atp] (wb) -- (wb_atp);
\node at (54.4625-3, 76.06-1) {\suwb};

\coordinate (wb_src_tgt) at (59.84, 76.02);
\coordinate (wb_src_tgt_atp) at (74.18, 78.07);
\draw [atp] (wb_src_tgt) -- (wb_src_tgt_atp);
\node at (59.84+7, 76.02-2) {\suwbst};

\addplot[only marks, g-blue, mark size=3] coordinates {
      (47.25, 67.91)
      (58.3075, 72.08)
      (54.4625, 76.06)
      (67.54, 77.44)
      (59.84, 76.02)
      (74.18, 78.07)
    };

\end{axis}
\end{tikzpicture}
\end{subfigure}
\caption{\small {\bf Left: (a)} Impact of using Domain Adaptive Pre-training. We fine-tune all the models on the source labeled data after pretraining. {\bf Right: (b)} Relationship between MLM (Masked LM) accuracy of pre-trained model and Entity-Linking performance of the fine-tuned model, evaluated on target domains. Adding domain adaptive pre-training improves
both MLM accuracy as well as the entity linking performance.
{\bf Note:} {\em src} represents
the union of all 8 training worlds and we adapt to one {\em tgt} world at a time. The target worlds are $\mcw_\text{tgt}^1$: {\em Coronation street}, $\mcw_\text{tgt}^2$: {\em Muppets}, $\mcw_\text{tgt}^3$: {\em Ice hockey}, $\mcw_\text{tgt}^4$: {\em Elder scrolls}.
$^\dagger$We refer to~\newcite{glorot2011domain}
for the idea of training a denoising autoencoder on source and target data together rather than the actual implementation. See text for more details.}
\label{fig_tab:adapt}
\vspace{-0.1in}
\end{figure*}

We re-implemented recent neural models designed for entity
linking~\cite{ganea-hofmann:2017:EMNLP2017, unseenroth}, but did not expect them to perform well since the original systems were designed for settings where labeled mentions or meta-data for the target entities were available.
The poor performance of these models validates
the necessity of using strong reading comprehension models for zero-shot entity linking.

When using the \mn model, pre-training is necessary to achieve reasonable performance.
We present results for models pre-trained on different subsets of our task corpus ($U_\text{src}$, $U_\text{tgt}$, \ust) as well as pre-training on an external large corpus (\uwb).
We observe that the choice of data used for pre-training is important.

In Table \ref{tab:baseline} we
also compare the Pool-Transformer, Candidate-Pool-Transformer and Full-Transformer.
The significant gap between \mn and the other variants shows the importance of allowing fine-grained comparisons between the two inputs via the cross attention mechanism embedded in the Transformer.
We hypothesize that prior entity linking systems did not need such powerful reading comprehension models due to the availability of strong additional meta information.
The remaining experiments in the paper use the Full-Transformer model, unless mentioned otherwise.

\subsection{Generalization to Unseen Entities and New Worlds}

To analyze the impact of unseen entities and domain shift in zero-shot entity linking, we evaluate performance on a more standard in-domain entity linking setting by making predictions on held out mentions from the training worlds.  Table~\ref{table:unseen} compares entity linking performance for different entity splits. Seen entities from the training worlds are unsurprisingly the easiest to link to. For unseen entities from the training world, we observe a 5-point drop in performance. Entities from new worlds (which are by definition unseen and are mentioned in out-of-domain text) prove to be the most difficult. Due to the shift in both the language distribution and entity sets, we observe a 11-point drop in performance. This large generalization gap demonstrates the importance of adaptation to new worlds.

\begin{table}[t]
\centering
\small
\begin{tabular}{l c}
\toprule
Evaluation & Accuracy \\
\midrule
Training worlds, seen & 87.74 \\
Training worlds, unseen & 82.96 \\
Validation worlds, unseen & 76.06 \\
\bottomrule
\end{tabular}
\caption{Performance of the Full-Transformer (\uwb) model evaluated on seen and unseen entities from the training and validation worlds. }
\label{table:unseen}
\vspace{-0.1in}
\end{table}

\subsection{Impact of Domain Adaptive Pre-training}
\label{sec:multistage}
Our experiments demonstrate that DAP improves on three state-of-the-art pre-training strategies:
\begin{itemize}
\setlength\itemsep{0.1em}
   \item {\bf \ust}: task-adaptive pre-training, which combines source and target data for pre-training~\cite{glorot2011domain}.\footnote{\footnotesize We use Masked LM and Transformer encoder, which are more powerful than the instantiation in {\cite{glorot2011domain}}.}
   \item {\bf \uwb}: open-corpus pre-training, which uses Wikipedia and the BookCorpus for pre-training (We use a pre-trained BERT model~\cite{devlin2018bert}).
   \item {\bf \uwbst}: the previous two strategies chained together. While no prior work has applied this approach to domain adaptation, a similar approach for task adaptation was proposed by \newcite{howard-ruder:2018}.
\end{itemize}

The results are in Figure~\ref{fig_tab:adapt}(a). 
DAP improves all pre-training strategies with an additional pre-training stage on only target-domain data. The best setting, \uwbstt, chains together all existing strategies. 
DAP improves the performance over a strong pre-trained model~\cite{devlin2018bert} by 2\%.

To further analyze the results of DAP, we plot the relationships between the accuracy of Masked LM (MLM accuracy) on target unlabeled data and the final target normalized accuracy (after fine-tuning on the source labeled data) in Figure~\ref{fig_tab:adapt}(b). 
Adding an additional pre-training stage on the target unlabeled data unsurprisingly improves the MLM accuracy. More interestingly, we find that improvements in MLM accuracy are consistently followed by improvements in entity linking accuracy.
It is intuitive that performance on unsupervised objectives reflect the quality of learned representations and correlate well with downstream performance. We show empirically that this trend holds for a variety of pre-training strategies.

\begin{table}[!t]
\centering
\small
\begin{tabular}{lcc }
\toprule
Pre-training & \multicolumn{2}{c}{EL Accuracy} \\
& N. Acc. & U. Acc. \\

\midrule 
\suwb{\footnotesize~\cite{devlin2018bert}}  & 75.06 & 55.08 \\
\midrule
\suwbtgt (DAP) &  76.17 & 55.88 \\
\suwbstt (DAP) & {\bf 77.05} & {\bf 56.58} \\

\bottomrule
\end{tabular}
\caption{Performance on test domains with \mn. {\bf N. Acc} represents the normalized accuracy. {\bf U. Acc} represents the unnormalized accuracy. The unnormalized accuracy is upper-bounded by 68\%, the top-64 recall of the candidate generation stage.}
\label{table:test_results}
\end{table}

\subsection{Test results and performance analysis}

Table~\ref{table:test_results} shows the normalized and unnormalized Entity Linking performance on test worlds. 
Our best model that chains together all pre-training strategies achieves normalized accuracy of 77.05\% and unnormalized accuracy of 56.58\%.
Note that the unnormalized accuracy corresponds to identifying the correct entity from tens of thousands of candidate entities. 

To analyze the mistakes made by the model, we compare EL accuracy across different mention categories in Table \ref{tab:categories}.
Candidate generation (Recall@64) is poor in the Low Overlap category.
However, the ranking model performs in par with other hard categories for these mentions.
Overall EL accuracy can thus be improved significantly by strengthening candidate generation.

\begin{table}[t!]
\small
\centering
\begin{tabular}{lccc}
\toprule
Mention Category & Recall@64 & \multicolumn{2}{c}{EL Accuracy} \\
& & N. Acc. & U. Acc. \\
\midrule
High Overlap &  99.28	& 87.64	& 87.00 \\
Ambiguous Substring &  88.03	& 75.89	& 66.81 \\
Multiple categories &  84.88	& 77.27	& 65.59 \\
Low Overlap &  54.37	& 71.46	& 38.85 \\
\bottomrule
\end{tabular}
\caption{Performance on test domains categorized by mention categories. Recall@64 indicates top-64 performance of candidate generation.
N. Acc. and U. Acc. are respectively the normalized and unnormalized accuracies.}
\label{tab:categories}
\end{table}

\section{Related Work}
We discussed prior entity linking task definitions and compared them to our task in section \ref{sec:taskdef}. 
Here, we briefly overview related entity linking models and unsupervised domain adaptation methods.

\paragraph{Entity linking models}
Entity linking given mention boundaries as input can be broken into the tasks of candidate generation and  candidate ranking. 
When frequency information or alias tables are unavailable, prior work has used measures of similarity of the mention string to entity names for candidate generation~\cite{anydb,hiermccallum}.
For candidate ranking, recent work employed distributed representations of mentions in context and entity candidates and neural models to score their compatibility. Mentions in context have been represented using e.g., CNN~\cite{hiermccallum}, LSTM ~\cite{unseenroth}, or bag-of-word embeddings~\cite{ganea-hofmann:2017:EMNLP2017}. Entity descriptions have been represented using similar architectures. To the best of our knowledge, while some models allow for cross-attention between single-vector entity embeddings and mention-in-context token representations, no prior works have used full cross-attention between mention+context and entity descriptions.

Prior work on entity linking tasks most similar to ours used a linear model comparing a mention in context to an entity description and associated structured data~\cite{anydb}. \newcite{anydb} also proposed a distant supervision approach which could use first-pass predictions for mentions in the target domain as noisy supervision for re-training an in-domain model. We believe this approach is complementary to unsupervised representation learning and could bring additional benefits. In another task similar to ours, \newcite{hengji} used collective inference and target database relations to obtain good performance without (domain, target database)-specific labeled training data. Collective inference is another promising direction, but could have limited success when no metadata is available.

\paragraph{Unsupervised domain adaptation} There is a large body of work on methods for unsupervised domain adaptation, where a labeled training set is available for a source domain and unlabeled data is available for the target domain.
The majority of work in this direction assume that training and test examples consist of $(x,y)$ pairs, where $y$ is in a fixed shared label set ${\cal Y}$. This assumption holds for classification and sequence labeling, but not for zero-shot entity linking, since the source and target domains have disjoint labels.

Most state-of-the-art methods learn non-linear shared representations of source and target domain instances, through denoising training objectives \cite{nlpJE}. In Section \ref{sec:adapt}, we overviewed such work and proposed an improved domain adaptive pre-training method.

Adversarial training methods~\cite{ganin2016domain}, which have also been applied to  tasks where the  space ${\cal Y}$ is not shared between source and target domains~\cite{cohen2018cross}, and multi-source domain adaptation methods~\cite{advmulti,reginamulti} are complementary to our work and can contribute to higher performance.

\eat{

\begin{itemize}[leftmargin=*]

\item Representation learning 
\begin{itemize}
\item Unsupervised pre-training \cite{glorot2011domain, chen2012marginalized}
\item Shared-Private representation learning (Frustratingly easy DA, domain separation networks)
\end{itemize}
\item Single/Multi-source domain adaptation via feature-space alignment
\begin{itemize}
\item DANN \cite{ganin2016domain}
\item MDAN \cite{zhao2018adversarial}
\item MAN \cite{chen2018multinomial}
\end{itemize}
\item Mixture-of-Experts for multi-source adaptation \cite{guo2018multi}
\item Cross-domain ranking \cite{cohen2018cross}
\item Domain-adaptive pre-training \cite{clinchant2016domain}
\item Unsupervised DA for entity linking
\end{itemize}
}

\eat{
In this work we focus on representation learning through denoising objectives and show how the best approach to learn such representations changes when a large non-task related corpus is used in addition to smaller task-related corpora. Our results show that a simple and general pre-training method outperforms the widely used approach of training on the union of source and target domains.

We expect that adversarial, shared-private, and multi-source domain adaptation methods could achieve additional complementary improvements.
}

\section{Conclusion}
We introduce a new task for zero-shot entity linking, and construct a multi-world dataset for it. 
The dataset can be used as a shared benchmark for entity linking research focused on specialized domains where labeled mentions are not available, and entities are defined through descriptions alone. 
A strong baseline is proposed by combining powerful neural reading comprehension with domain-adaptive pre-training.

Future variations of the task could incorporate NIL recognition and mention detection (instead of mention boundaries being provided). 
The candidate generation phase leaves significant room for improvement.
We also expect models that jointly resolve mentions in a document would perform better than resolving them in isolation.

\section*{Acknowledgements}
We thank Rahul Gupta and William Cohen for providing detailed helpful feedback on an earlier draft of this paper.
We thank the Google AI Language Team for valuable suggestions and feedback.

\bibliography{main}
\bibliographystyle{acl_natbib}

\appendix

\section{Examining model errors and predictions}
In tables \ref{e1}, \ref{e2}, \ref{e3}, \ref{e4} we show some example mentions and model predictions. 
For each instance, the examples show the correct gold entity and the top-5 predictions from the model. 
Examples show 32 token contexts centered around mentions and the first 32 tokens of candidate entity documents.

\onecolumn

\begin{table*}[!t]
\centering
\footnotesize
\begin{tabularx}{\textwidth}{c L{2.6cm} X}
\toprule
\multicolumn{3}{c}{\bf Coronation Street} \\
\midrule

\textit{Mention} &  \multicolumn{2}{p{0.85\textwidth}}{
Robbie pulled over the ambulance with a van and used a gun to get the 
{\textbf{Prison Officer}}
with Tony to release him . He integrated himself with the Street residents , finding}\\

\midrule

\textit{Gold Entity} & 
Prison Officer (Episode 7351)	&
The unnamed Prison Officer was on duty during May 2010 in the Highfield Prison dining room when Tony Gordon provoked a fight with a fellow inmate \\

\midrule 
\multicolumn{3}{c}{\textit{Top-5 predictions}} \\
\midrule 

\checkmark & Prison Officer (Episode 7351) & 
The unnamed Prison Officer was on duty during May 2010 in the Highfield Prison dining room when Tony Gordon provoked a fight with a fellow inmate \\\\

& Inmate (Episode 7351) & 
The Inmate was an unnamed fellow prisoner of Tony Gordon in Highfield Prison . Tony provoked a fight in the dining room with the inmate by staring \\\\

& Police Officer (Simon Willmont)	&
The unnamed Police Officer was on duty at Weatherfield Police Station in March 2010 when Peter Barlow was released from custody following his arrest as he \\\\

& Prison Officer (Bill Armstrong)	& 
The Prison Officer looked after the incarceration of three Coronation Street residents : In November 2000 he was on duty at Strangeways Jail when Jim McDonald \\\\

& Robbie Sloane	&
Quietly spoken Robbie Sloane was Tony Gordon ' s henchman and a convicted murderer , who he met while sharing a cell at Highfield Prison in 2010 . When Robbie \\
\bottomrule 
\end{tabularx}
\caption{Mention and entity candidates from Coronation Street.}
\label{e1}
\end{table*}

\begin{table*}[!t]
\centering
\footnotesize
\begin{tabularx}{\textwidth}{c L{2.6cm} X}

\toprule
\multicolumn{3}{c}{\bf Muppets} \\
\midrule

\textit{Mention} &  \multicolumn{2}{p{0.85\textwidth}}{
Bean Bunny was introduced during the seventh season of " Muppet Babies " , and a
{\textbf{pre - teen Bean}}
would later be featured as part of the Muppet Kids series . Bean was active}\\ 

\midrule

\textit{Gold Entity} & 

Bean Bunny (Muppet Kids) &
A young version of Bean Bunny made a few appearances in the Muppet Kids books and video games . Young Bean moves to the Muppet Kids\\\\

\midrule 
\multicolumn{3}{c}{\textit{Top-5 predictions}} \\
\midrule 

& Baby Bean Bunny &
Baby Bean Bunny appeared in the late 1989 / 1990 seasons of " Muppet Babies " as a baby version of Bean Bunny . He joined the other babies\\\\

\checkmark & Bean Bunny (Muppet Kids) &
A young version of Bean Bunny made a few appearances in the Muppet Kids books and video games . Young Bean moves to the Muppet Kids\\\\

& Bean Bunny &
Bean Bunny first appeared in 1986 as the star of the TV special " The Tale of the Bunny Picnic " . The cute bunny was part of a family\\\\

& Piggy (Muppet Kids) & 
A pre - teen version of Miss Piggy , as seen in the " Muppet Kids " books and video games . Piggy lives in a fancy\\\\

& Muppet Kids & 
Muppet Kids was a series of books and educational software made in the 1990s , featuring young , pre - teen versions of the principal franchise characters . Characters included\\

\bottomrule 
\end{tabularx}
\caption{Mention and entity candidates from Muppets.}
\label{e2}
\end{table*}

\newpage

\begin{table*}[!t]
\centering
\footnotesize
\begin{tabularx}{\textwidth}{c L{3.2cm} X}
\toprule

\multicolumn{3}{c}{\bf Ice Hockey} \\
\midrule

\textit{Mention} &  \multicolumn{2}{p{0.85\textwidth}}{
1979 - 80 PCJHL Season This is a list of
{\textbf{Peace - Cariboo Junior Hockey League}}
Standings for the 1979 - 80 season . This was the PCJHL ' s final}\\

\midrule

\textit{Gold Entity} & 

Rocky Mountain Junior Hockey League &
The Rocky Mountain Junior Hockey League was a Canadian Junior " A " ice hockey league in British Columbia . History . Promoted to a Junior " \\

\midrule 
\multicolumn{3}{c}{\textit{Top-5 predictions}} \\
\midrule 

& Peace Junior Hockey League &
Hockey League Peace Junior Hockey League is a League that started in the 1960 ' s and ended in 1975 . Then change its name to Peace Cariboo junior Hockey\\\\

& Cariboo Hockey League &
The Cariboo Hockey League was a Senior and Intermediate hockey league in the Cariboo District of British Columbia , Canada . History . The league began in the 1955\\\\

& Cariboo Junior League &
The Cariboo Junior League operated in northern British Columbia in the 1963 - 64 season . Its champion was eligible for the British Columbia Junior Playoffs . The league\\\\

\checkmark & Rocky Mountain Junior Hockey League &
The Rocky Mountain Junior Hockey League was a Canadian Junior " A " ice hockey league in British Columbia . History . Promoted to a Junior "\\\\

& North West Junior Hockey League &
The North West Junior Hockey League is a Junior " B " ice hockey league operating in the Peace River region of Alberta and British Columbia ,\\

\bottomrule 
\end{tabularx}
\caption{Mention and entity candidates from Ice Hockey.}
\label{e3}
\end{table*}

\begin{table*}[!ht]
\centering
\footnotesize
\begin{tabularx}{\textwidth}{c L{3.2cm} X}
\toprule

\multicolumn{3}{c}{\bf Elder Scrolls} \\
\midrule

\textit{Mention} &  \multicolumn{2}{p{0.85\textwidth}}{
to get everyone to safety . Rolunda ' s brother is one of those people .
{\textbf{The Frozen Man}}
. Rolunda ' s brother Eiman has ventured into Orkey ' s Hollow to find}\\

\midrule

\textit{Gold Entity} & 

The Frozen Man (Quest) &
The Frozen Man is a quest available in The Elder Scrolls Online. It involves finding a Nord who has been trapped in ice by a mysterious " Frozen Man \\

\midrule 
\multicolumn{3}{c}{\textit{Top-5 predictions}} \\
\midrule 

\checkmark & The Frozen Man (Quest) &
The Frozen Man is a quest available in The Elder Scrolls Online. It involves finding a Nord who has been trapped in ice by a mysterious " Frozen Man\\\\

& The Frozen Man & 
The Frozen Man is an insane Bosmer ghost found in Orkey ' s Hollow . He says he was in a group of people inside the cave when it\\\\

& Kewan	& 
Kewan is a Redguard worshipper of the Daedric Prince Peryite . He is frozen in a trance that relates to the Daedric quest , but can be unfrozen in completion the \\\\

& Stromgruf the Steady & 
Stromgruf the Steady is the Nord who is found in the Grazelands of Vvardenfell , west of Pulk and east of Vassamsi Grotto ( Online ) . He is \\\\

& Maren the Seal & 
Maren the Seal is a Nord hunter and worshipper of the Daedric Prince Peryite . She is frozen in a trance that relates to the Daedric Prince ' s\\\\

\bottomrule 
\end{tabularx}
\caption{Mention and entity candidates from Elder Scrolls.}
\label{e4}
\end{table*}

\end{document}